\documentclass[a4paper, 10pt, twocolumn, superscriptaddress, nofootinbib, showkeys, notitlepage]{revtex4-1}
\usepackage[english]{babel}
\usepackage{amsmath, amssymb, amsfonts, amscd, amsthm}
\usepackage[a4paper,top=3.5cm,bottom=3cm,left=2.5cm,right=2.5cm,bindingoffset=5mm]{geometry}
\usepackage{dsfont}
\usepackage{mathrsfs}
\usepackage{stmaryrd}
\usepackage[utf8x]{inputenc}
\usepackage{indentfirst}
\usepackage{microtype}
\usepackage{graphicx}%insert figures
\usepackage{float}
\usepackage[margin=.5cm, labelfont=it]{caption}
\usepackage{subcaption}
\usepackage{textcomp, enumitem}
\usepackage{array, booktabs}
\usepackage[colorlinks=true, linkcolor = blue, citecolor= red, urlcolor= red]{hyperref}
\usepackage{tikz, tikz-cd}
\usepackage{tcolorbox}
\usetikzlibrary{arrows, calc, shapes, fit, matrix, decorations.pathreplacing, decorations.markings, decorations.pathmorphing, positioning, fadings, patterns, chains, matrix, backgrounds}
\usepackage[all]{xy}
\usepackage{stackengine,scalerel}
\usepackage{mathtools}
\usepackage{numprint}
\npthousandsep{\,}
\usepackage{cleveref}

\tikzstyle{vecArrow} = [thick, decoration={markings,mark=at position
   1 with {\arrow[semithick]{open triangle 60}}},
   double distance=1.4pt, shorten >= 5.5pt,
   preaction = {decorate},
   postaction = {draw,line width=1.4pt, white,shorten >= 4.5pt}]
\tikzstyle{innerWhite} = [semithick, white,line width=1.4pt, shorten >= 4.5pt]

\begin{document}

%%%%%%Authors details%%%%%%

\author{Oscar de Felice}
	\affiliation{Sorbonne Universités UPMC Paris 06, CNRS, UMR 7589, LPTHE, F-75005, Paris, France\vspace{2ex}}
	{\let\thefootnote\relax\footnote{\hspace{-4ex}\textsuperscript{1}\emph{e-mail}: \href{mailto:oscar.defelice@gmail.com}{odefelice@lpthe.jussieu.fr}}}
%%%
\author{Gustavo de Felice}
	\affiliation{via Silvio Pellico 12, 84073 Sapri, Italy}
	{\let\thefootnote\relax\footnote{\hspace{-4ex}\textsuperscript{2}\emph{e-mail}: \href{mailto:gustavo@dottordefelice.it}{gustavo@dottordefelice.it}}}
%%%

%%%%
%%%%%%%%%%%%%%%%%%%%

\date{\today}
%
%%
%%%
%%
%
%
%%%%%%Title%%%%%%%%%%%%
\title{%
	 A smile I could recognise in a thousand \\[1.5mm]
  	\small Automatic identification of identity from dental radiography
	}
%%%%%%%%%%%%%%%%%%%%
%
%%%%%%%%%%%%
\begin{abstract}
	In this paper, we present a method to automatically compare multiple radiographs in order to find the identity of a patient out of the dental features.
	The method is based on the matching of image features, previously extracted by computer vision algorithms for image descriptor recognition.
	The principal application -- being also our motivation to study the problem -- of such a method would be in victim identification in mass disasters.
\end{abstract}
%%%%%%%%%%%%%
%
\maketitle

\section*{Introduction}
	An important issue to face since the birth of image recognition techniques is the detection and classification of objects in digital images. 
	Obviously, objects can be classified by several aspects, \emph{e.g.} colours, textures, shapes, position within images, etc.
	A fertile application of image recognition methods are computer aided medical diagnostic systems, whose first example dates back to 1963~\cite{lodwick1963computer}.
	
	Recently, there have been several satisfying examples of the use of such approach~\cite{castanon2007, Arena2003, Ushizima2004}. 
	Even back to the early 2000's, one can find contributions to the applications image recognition to medical diagnostic systems~\cite{Comaniciu1999, Jain2000}.
	
	A nice and complete review of the argument can be found in~\cite{cadReview} and references in there.
	
	Another important field of application of image recognition, even if less deeply explored, is the forensic medicine.
	Indeed, here the interest to the particular subject rather than to the general features of diseases does not make immediate the use of statistical methods, hence machine learning.
	
	Nowadays, image recognition algorithms have become quite efficient, thanks to the availability of more powerful and less expensive machines.
	This allowed the applications of automatic methods to face several issues. 
	For instance fingerprint recognition~\cite{Ratha:2003}, age estimation~\cite{cameriere:2015, bacaloni}, pregnancy screening~\cite{pregnScreen} and many others.
	
	The authors of~\cite{cameriere:2015} in particular, make use of teeth radiographs to estimate the age of patients out of images. 
	More specifically, it is explained how to use periapical radiographs of canines to apply the so-called \emph{Cameriere's method} for age estimation~\cite{cameriere:2004, cameriere:2006, cameriere:2008}.
	
	This paper aims to explore the possibility of subject identification by an algorithm of dental features matching.

	%%%
	%%
	%
\subsection*{Motivations}
	A motivational example might be the following: imagine the case of a \emph{mass disaster}\footnote{%
		A mass disaster may be defined as a man-made or natural disaster in which so many persons are injured that local emergency medical services are overwhelmed and/or destroyed.%
		}. 
	In such an event it is crucial the precise and rapid identification of victims, a process that can be quite complicate.
	In forensic medicine, there are usually three methods to identify corpses, said \emph{primary methods}.
	These are fingerprints comparison, DNA analysis and teeth shape matches. 
	In the case of a mass disaster, fingerprints might be compromised, while DNA analysis can be slow and expensive, thus often teeth study is the most efficient way to identify corpses.
	An image comparison system might help the forensic odontologist to compare hundreds of radiographs and find the one with highest similarity score with the given one.
	
	Furthermore, a model like this might be useful for many other reasons: \emph{e.g.} again age estimation from dental radiographs or it might help dentists to study the evolution of a disease, etc.
	%
	
	%%%
	%%
	%
\subsection*{Content of the paper}
	Here we would like to perform a comparison between images by feature matching.
	In order to compare two images, one has to identify the characteristic objects of them.
	This procedure is known under the name of \emph{feature detection}.
	There are several methods to detect features in an image. 
	In our code we implemented three of them, \textsc{sift} (Scale-Invariant Feature Transform)~\cite{lowe1999sift, lowe2004sift}, \textsc{surf} (Speeded Up Robust Feature)~\cite{surf:2006} and \textsc{orb} (Oriented FAST and Rotated BRIEF)~\cite{orb} that can be used according to the one giving the highest comparison score.
	This because of the remarkable performances of these algorithms in comparison with other descriptors~\cite{performanceSift}.
	We describe such a procedure in~\ref{sec:SIFT}.
	Indeed, to define a score we need to define a \emph{metric}, measuring how much two images are similar.
	This is done in section~\ref{sec:metr}, where we give the details about what we called \emph{Lowe distance}, after David Lowe, who invented \textsc{sift} algorithm~\cite{lowe1999sift}.
	Finally, we give details about feature detection of an image in the appendix~\ref{app:SIFT}.
	
	%
	%
	%%%%%
	%%%
	%%
	%
\section{Materials}
	\label{sec:materials}
	This preliminary section has the role of describing the images we are going to use in order to build and apply our method.
	
	Radiographs come anonymously from the database of dr Gustavo de Felice.
	The identity of patients is hidden by a code whose meaning is unknown to the authors of this article.
	Although this, the database owner is also in possession of signed authorisation forms, allowing the use of the images for research purposes.
	%
	%
	%%%%%
	%%%
	%%
	%
\section{Methods}
	\label{sec:methods}
	As mentioned above, we start by selecting an image whose owner code is known.
	Here and in the following, we call this \emph{test image}.
	This would correspond to the patient to be identified in the case of a mass disaster.

	Hence, the whole corpus of the remaining images is called \emph{training set} following name convention of Machine Learning problems.
	
	The procedure is quite simple: selected the test image, we calculate its similarity score (see section~\ref{sec:metr}) with all the images in the training set.
	The image with highest similarity score is our candidate to be the corresponding one.
	A careful reader may wonder what would happen if in the test set there is no other image of the patient to identify.
	To avoid the case of a wrong identification, we set a similarity threshold, such that if no image in the test set has similarity score -- with the train image -- lower than the threshold, then the procedure will exit without any identification.
	%
	%%%%%
	%%%
	%%
	%
\section{Algorithms for keypoint localisation}
	\label{sec:SIFT}
	In this section we review the general scheme of algorithms for feature identification and matching. 
	We have devoted an appendix to describe one of the most used algorithms, \emph{i.e.} \textsc{sift} which stands for  \emph{Scale Invariant Feature Transform}, hence here we only illustrate the general approach, while relegating further details to appendix~\ref{app:SIFT}.
	Thus, we are going to expose the working mechanism of feature recognition algorithm and its application to teeth image comparisons.
	
	To begin, let's describe the feature recognition part of the algorithm.
	%
	%
	%%%
	%%
	%
\subsection{Feature localisation}
	Mathematically, images can be thought as sets of connected points in a two-dimensional space, often (and also here) approximated in a discrete binary space.
	People do not perform image classification directly on, since this task is computationally really expensive ($\mathcal{O}(n^2)$), where each image is made up by $n$ pixels.
	The representation of an image can be modified by an image transformation, by mapping the space $\tilde{\mathcal{F}}$ to a -- typically smaller -- feature space $\mathcal{F}$.
	
	The features we want to localise are image points endowed with the right properties under some symmetries.
	Indeed, we are looking for something scale invariant, colour-space independent, rotational and translational invariant.
	
	All the algorithms share the same philosophy: we find some candidate keypoints and then we cut off the ones spoiling the invariance, landing on the final set.
	In the case of \textsc{sift} the keypoints are localised in a scale space built on the feature space $\mathcal{F}$ by a convolution with gaussians.
	%
	%
	%%%
	%%
	%
\subsection{Feature matching}
	Once the images have been endowed with their keypoint descriptors, we have to match the descriptors of different images.
	There are several methods to match objects which are represented numerically by arrays.
	In our program we implemented the simplest one: the brute-force method, where we measure a distance between matching descriptors of two images.
	In order to accept or refuse a match we make use of the so-called \emph{Lowe ratio method}.

		\begin{figure}[!ht]
			\centering
			\includegraphics[width=\linewidth]{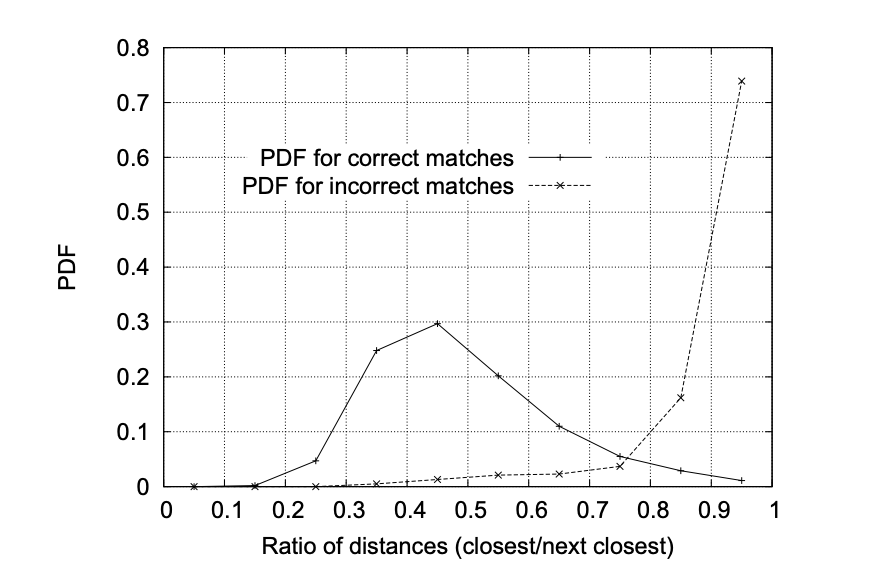}
			\caption{	Image shows the probability distribution of relative distances between keypoints.
					The solid lines represents how good matches have a $0.45$ centred distribution, while bad matches are localised at great distances, as one might expect.
					For these reasons, statistically, fixing a threshold of $0.7$ we have a good confidence of getting most of good matches.
					Image comes from~\cite{lowe2004sift}.}
		\label{fig:LoweRatio}
		\end{figure}

	Lowe ratio test works as follows. 
	One can compare the distance of the best match, with the second best one. 
	Taking into account distances between good matches and bad ones, one can plot a probability distribution as the one in figure~\ref{fig:LoweRatio}.
	
	Such a measure is based on the observation a good match between two keypoints $k_A$ and $k_B$ for images $A$ and $B$ will have a distance (or a score) really different with respect to its closest matches.
	In other word, given a good match $(k_A, k_B)$ and its distance $\delta(k_A, k_B)$, we state that the match $(k_A, k'_B)$ has a much greater distance, while the distance difference for two bad matches is small in comparison to characteristic match scale. 
	
	To rephrase, a good match will be sourrounded by bad matches, and the ratio 
		\begin{equation*}
			\mathscr{L} = \frac{\delta(k_A, k_B)}{\delta(k_A, k'_B)}\, ,
		\end{equation*}
	will be closer to zero the more the match is good, on the opposite it will be close to $1$.
	We fix a threshold for $\mathscr{L}$ at $70\%$, following~\cite{lowe2004sift, thressift}.
	
	%
	%
	%%%%%
	%%%
	%%
	%
\section{Applying feature identification to radiographs}
	The aim of this section is to go in depth and describe how the algorithm works.
	However, before digging into details, we want to give an introductory scheme, playing the role of a summary.
	
	As described above the training set is composed by a large set of images. 
	In our motivational example, these are the dental radiographs of victims of a mass disaster.
	The test image is a radiograph of someone whose identity is known and we would like to verify whether he is amongst the victims.
	
	We are going to apply feature detection to all images, and then by keypoint comparison and ratio test we are going to find the best correspondence for the image.
	A threshold for the metric has been defined, such that if the test subject is not in the victims set, we do not make bad identifications.
	Scores of all image couples are calculated.
	We take all the scores and rescale them on a Gaussian. 
	We label the correspondence as good if and only if its rescaled score is over the $66\%$, otherwise we conclude the test subject was not amongst victims.
	
	%
	%%%
	%%
	%
\subsection{Metric definition}
	\label{sec:metr}
	The goal of this section is to define a \emph{metric} on the space of images.
	Technically speaking, we have a space of images $\tilde{\mathcal{F}}$ that we endow with the concept of distance.
	
	Recall a distance $\delta$ must satisfy the following properties
		\begin{align*}
			\delta(x, y)& \geq 0 & \forall x, y\, ,\\
			\delta(x, y)& = 0 \Leftrightarrow x = y\, , & \forall x, y\, ,\\
			\delta(x, y)& = \delta(y,x)\, , & \forall x, y\, ,\\
			\delta(x, y)& \leq \delta(x,z) + \delta(z, y) & \forall x, y, z\, .
		\end{align*}
	One can see, for instance~\cite{topologybook} for details.
	
\begin{figure}[htb]
   	 	\centering
			\begin{subfigure}{0.2\textwidth}
  				\includegraphics[width=\linewidth]{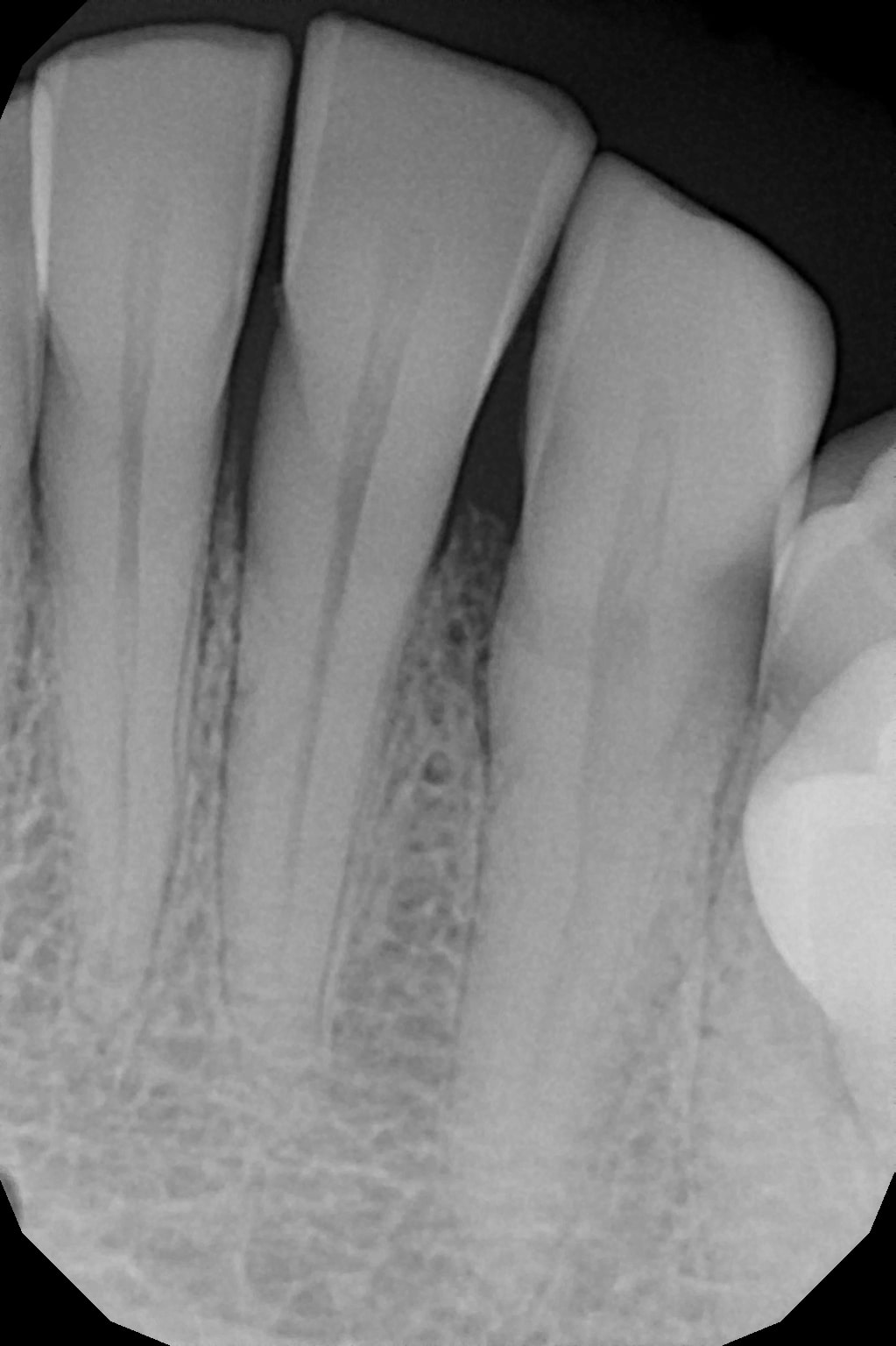}
  				\caption{Radiograph $1$}
  				\label{rx:1}
			\end{subfigure}\hfil 
			\begin{subfigure}{0.2\textwidth}
  				\includegraphics[width=\linewidth]{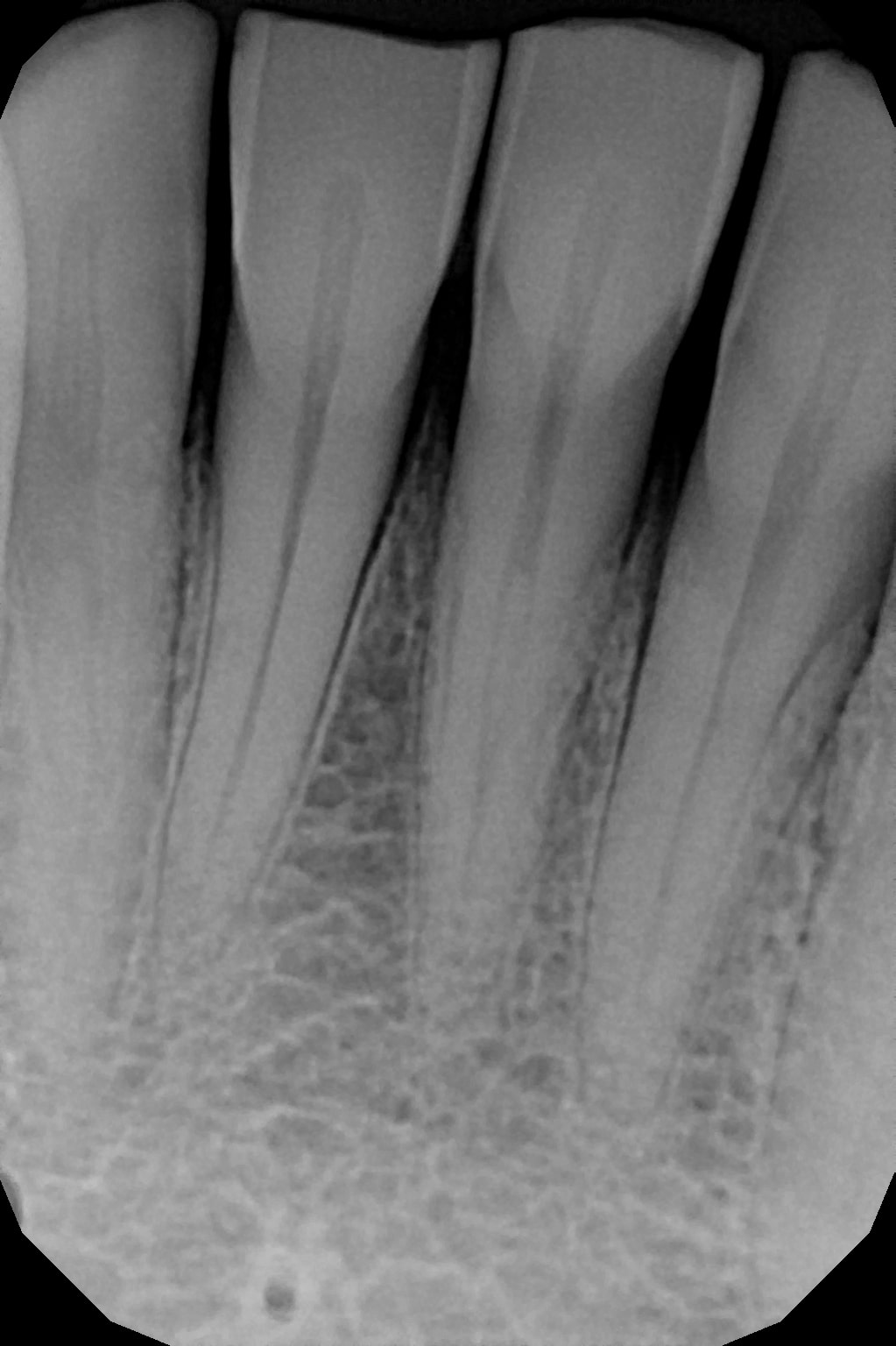}
  				\caption{Radiograph $2$}
  				\label{rx:2}
				\end{subfigure}\hfil
			\caption{How similar are these two images?}
			\label{fig:rx}
		\end{figure}

\begin{figure}[htb]
   	 	\centering
			\begin{subfigure}{0.2\textwidth}
  				\includegraphics[width=\linewidth]{Images/rx1.jpg}
  				\caption{Image $1$}
  				\label{fig:img1}
			\end{subfigure}\hfil 
			\begin{subfigure}{0.2\textwidth}
  				\includegraphics[width=\linewidth]{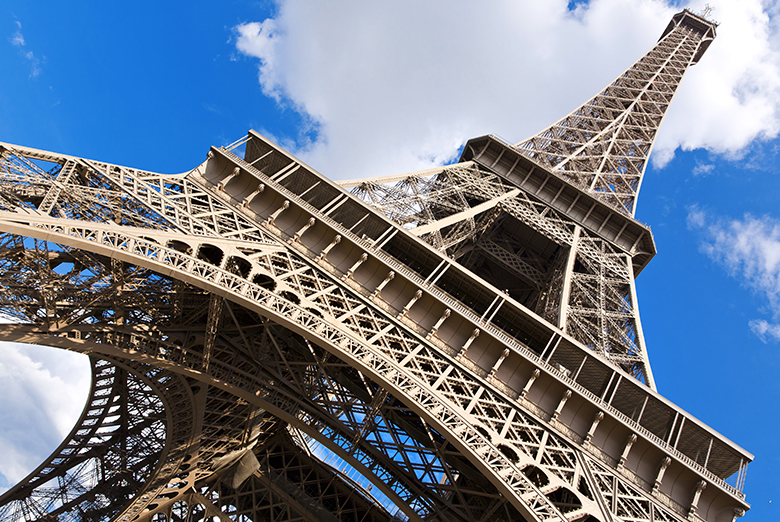}
  				\caption{Image $2$}
  				\label{fig:img2}
				\end{subfigure}\hfil
			\caption{How similar are these two images?}
			\label{fig:imgs}
		\end{figure}

	What we aim to do is to define a numerical measure -- satisfying the properties above -- to say whether two images are close or far and especially to say how much close they are.
	To be concrete, take into account figures~\ref{fig:rx} and~\ref{fig:imgs}. 
	It is clear that the figure~\ref{rx:1} is closer to~\ref{rx:2} than to~\ref{fig:img2}. 
	However, it is not immediately clear how to state this in a mathematical way. 
	We have to define how to give a score of similarity stating $\delta(\mathrm{rx}_1,\mathrm{rx}_2) < \delta(\mathrm{rx}_1,\mathrm{img}_2)$.
	
	We are going to define a function called \emph{Lowe~distance} modelling the distance between two images and based both on the number of features matching and on the score of matching.
	Given the definition we are going to prove it satisfies the properties above.

	The distance $\delta$ is defined by taking the Lowe ratio collecting good matches and getting the ratio between good matches and total matches.
	Explicitly, given two images $A$ and $B$. 
	We have  two sets of keypoints $K_A = \{k_a\}_{a \in A}$ and $K_B =\{k_b\}_{b \in B}$.
	The set of good matches is given by the keypoints passing the Lowe ratio test described above.
	We denote such a set as $\mathcal{K}(A,B)$.
	Having this, one can define the \emph{Lowe distance} between the two images as
		\begin{equation*}
			\mathcal{L}(A,B) := \frac{\#\mathcal{K}(A,B)}{\max(\#K_A, \# K_B)}\, .
		\end{equation*}

	It is on such a metric that we set the score of $66\%$ as stated above.
	
	%
	%
	%%%
	%%
	%
\subsection{Procedure}
\begin{figure}[!ht]
			\centering
			\begin{tikzpicture}
					% Test image
					\node[inner sep=0pt] (RX) at (2,0) {\includegraphics[width=.15\textwidth]{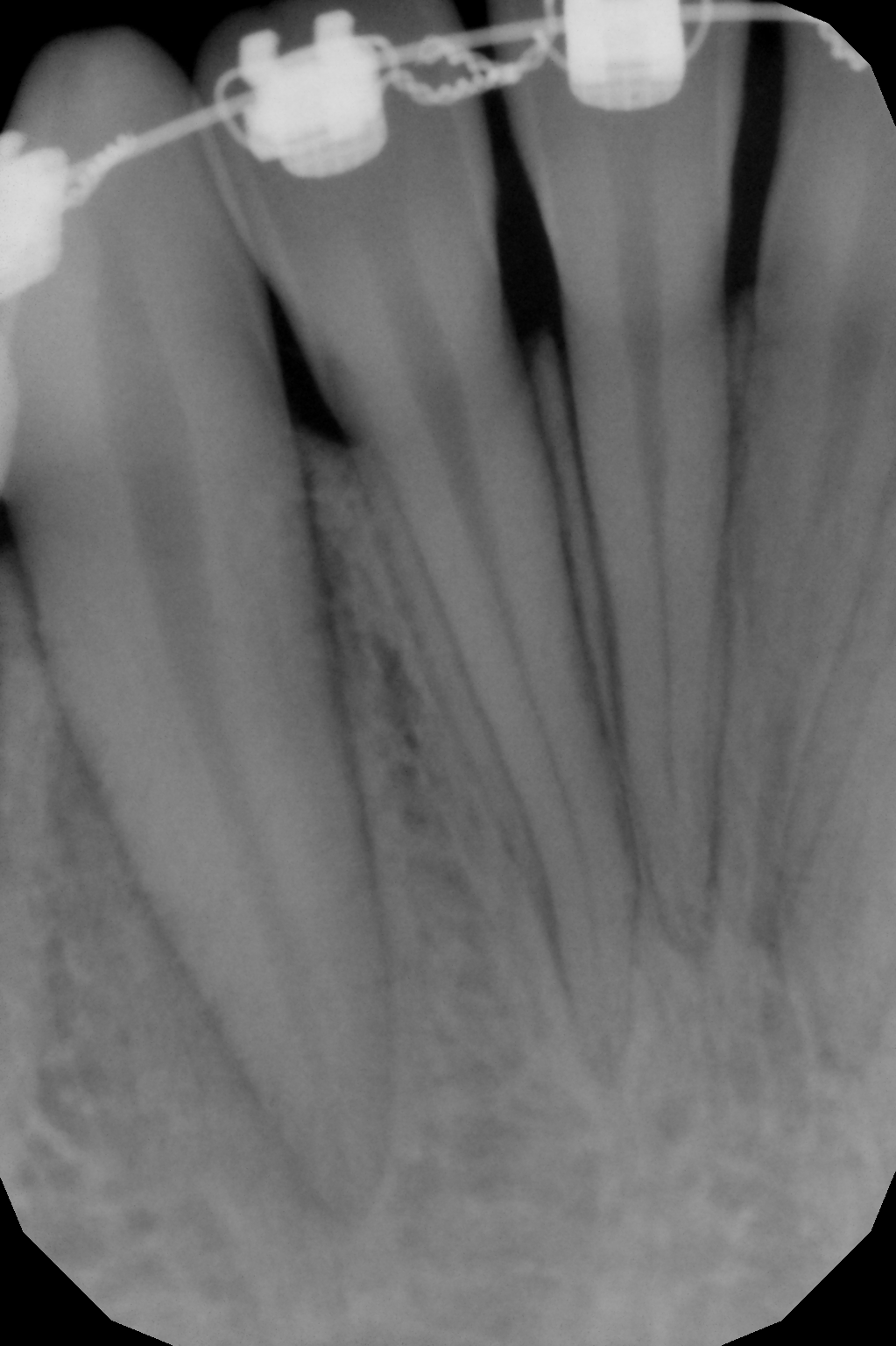}};
					% Train set of images
					\node[inner sep=0pt] (RX1) at (1.5,-6) {$\ldots$};
					\node[inner sep=0pt] (RX2) at (2.5,-6) {\includegraphics[width=.05\textwidth]{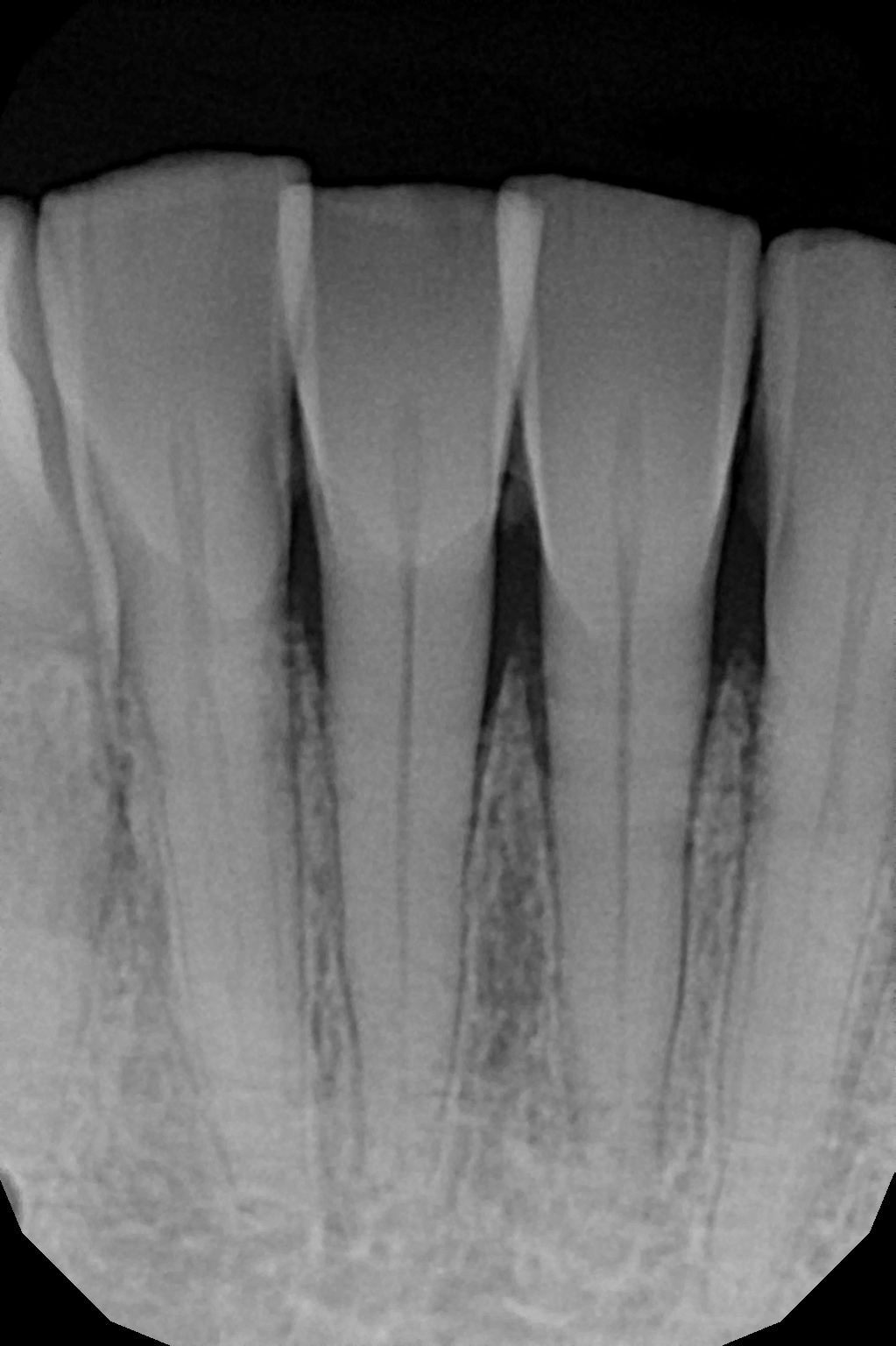}};
					\node[inner sep=0pt] (RX3) at (3.5,-6) {\includegraphics[width=.05\textwidth]{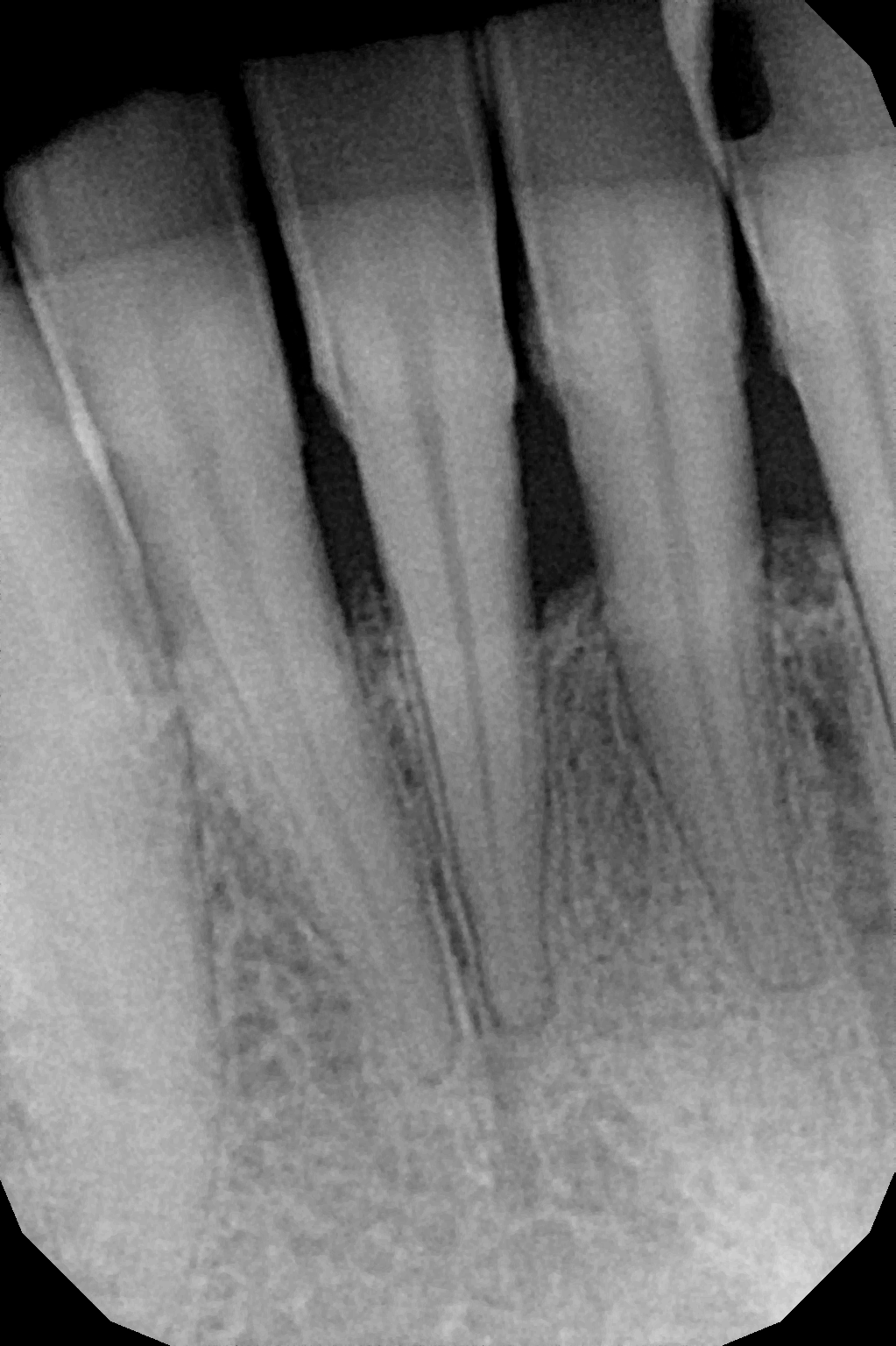}};
					\node[inner sep=0pt] (RX4) at (5.5,-6) {\includegraphics[width=.05\textwidth]{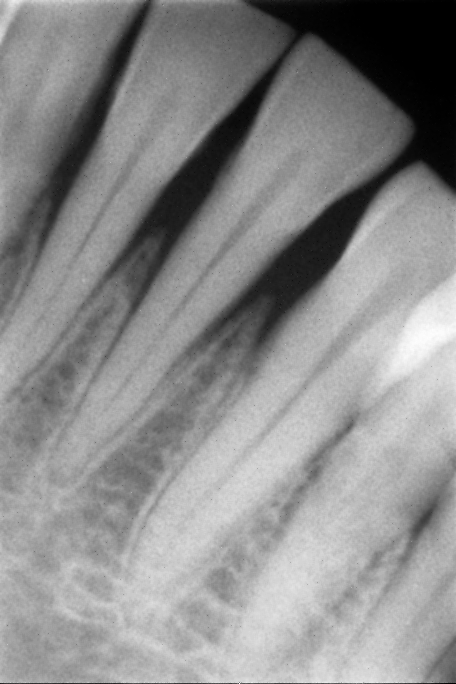}};
					\node[inner sep=0pt] (RXlabeled) at (4.5,-6) {\includegraphics[width=.05\textwidth]{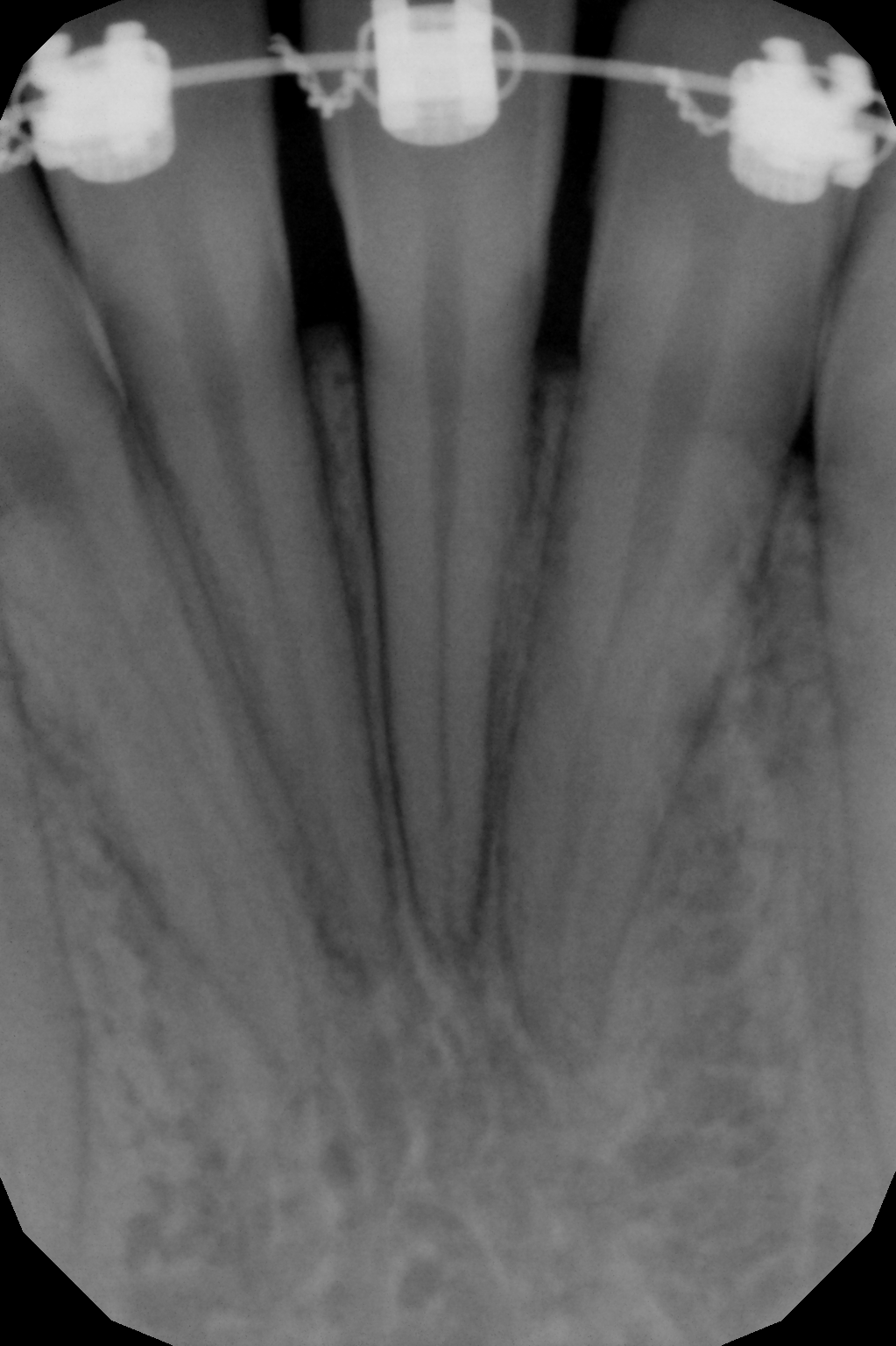}};
					\node[inner sep=0pt] (RX5) at (6.5,-6) {$\ldots$};
					% Arrow
					\draw[->] (RX.south east) to[out=-30, in=75, looseness=4, edge node={node[midway,fill=white, text width=5cm] {Has the test image a correspondence?}}] (RXlabeled.north);
					%\draw[->,thick] (RX.south east) -- (RXlabeled.north west)
		 %node[midway,fill=white, text width=5cm] {A quale immagine corrisponde l'immagine di \emph{test}?};
				 	% Rectangle
				 	\draw[red,thick,dotted] ($(RXlabeled.north west)+(-0.1,0.3)$) rectangle ($(RXlabeled.south east)+(0.1,-0.3)$);
			\end{tikzpicture}
			\caption{ The process of identification goes on by descriptors comparison.\\%
								Each image has been associated to a vector of descriptors.
								Numerically, we can compare vectors and compute a score for each match.
								The image in the train set, having the greatest score is chosen as the corresponding one.}
		\label{img:matching_series}
		\end{figure}

	The procedure is schematically illustrated in figure~\ref{img:matching_series}.

	To be more descriptive we are going to illustrate the various phases of the procedure.
	First of all, as explained above, we have the test image whose we know the identity, denoted by $\mathcal{I}_{\mathrm{test}}$.
	The training set is made up by a collection of images $\{ \mathcal{I}_{\mathrm{train}} \}$.
	We compute and identify the features of the test image, to have a set of descriptor vectors characterising the image.
	One can do the same on each image of the training set, having a set of descriptors for each image.
	At this stage, it is time to proceed to the comparison.
	There are several methods (we implemented brute force and \textsc{flann} matchers) to compare descriptors. 
	Conceptually, we stick on the brute force one for the sake of simplicity.
	
	Hence, we take the Euclidean distance between vectors, apply the Lowe ratio test to split matches in good and bad ones and select the best image for matching score.
	
	If the best matched image $\mathcal{I}^{\mathrm{best}}_{\mathrm{train}}$ has a score-ratio $\mathcal{L}(\mathcal{I}_{\mathrm{test}}, \mathcal{I}^{\mathrm{best}}_{\mathrm{train}})$ over the threshold value we set, then we say it has been identified. 
	Otherwise, we reject the identification, stating the test subject was not amongst the victims.
	
	%	
	%
	%%%
	%%
	%
\subsection{Discussion}
	There are few details we have been sloppy about in the procedure description above.
	
	We begin by saying that actually in computation we did not use the brute force method to compare descriptors, mainly for numerical reasons.
	Indeed, as showed in~\cite{flann}, a square root kernel instead of the standard Euclidean distance to measure the similarity between descriptors leads to a dramatic performance boost in all stages of the pipeline.
	
	We are not going into details about the kernel on which \textsc{flann} matcher is based, referring to~\cite{flann, ballard} for a detailed discussion.

	To go on, in the procedure described above the keypoint descriptors calculation was performed for each image and then we compared vectors and applied the Lowe ratio test.
	This gave us nice results, however, it is easy to understand this is not the most efficient way of comparison.
	
	Numerically, a smart ploy could be to collect the descriptors of an image with a clustering algorithm. 
	This is a way to perform a sort of feature reduction, by taking only, for example, the highest contrast keypoints, which are the ones giving more information about objects inside the image.
	
	%	
	%
	%%%
	%%
	%
\subsection{Results}
	This brief section to expose results of method described above.
	
	We performed identification through this procedure over $100$ test images, having a training set of $1554$ images. 
	The training set was actually composed by images whose owner code was known, but hidden to the algorithm.
	This in order to be able to check the predictions. 
	We got a correct identification in $99$ cases.
	We also tried to identify $50$ radiographs having no correspondence in the training set. 
	In this case results are less impressive, but still satisfying. 
	The number of correctly rejected identification is $46$.
	
	These results are collected in table~\ref{tab:results}.
			\begin{table}[!ht]
			\centering
				\begin{tabular}{cccc}
				\toprule
 								&	\textsc{sift}	& 	\textsc{surf}		& 	\textsc{orb}		\\
				\midrule
				\textsc{Precision}	&	$89\%$		&	\underline{$92\%$}	&	$90\%$			\\
				\textsc{Recall}		&	$98\%$ 		& 	\underline{$99\%$}	&	\underline{$99\%$}	\\
				\bottomrule
				\end{tabular}
			\caption{	Rates of identification.
					We expressed results for the three implemented feature extraction algorithms in terms of the well-known measures of precision and recall.}
			\label{tab:results}
			\end{table}
		%

	%
	%
	%%%%%
	%%%
	%%
	%
\section{Perspectives and Conclusions}
	This paper describes the first proposal for a computer aided identity recognition system based on dental images.
	The discussion goes through different aspects in describing the implemented method.
	
	First, we describe the key motivational example: corps identification through dental features in mass disasters.
	Then, we skew the algorithm working scheme.
	After that, we formally define a metric, allowing us to measure how close are two images.
	This gives us also a numerical value estimating how reliable is the identification that comes out of the procedure.
	
	Finally, we discuss the possible caveats of the method and try to propose some solutions.
	
	To conclude, this paper illustrates an example of how it is possible to work at the interface between various fields.
	There are quite a lot of examples of computer vision applications to diagnostics, see for instance~\cite{Arena2003, Comaniciu1999, Ushizima2004, castanon2007, lodwick1963computer} and references in there.
	For a more complete review of the subject, one can look at~\cite{cadReview}.
	
	Here, artificial intelligence, forensic medicine and computer science are involved.
	The already cited works~\cite{bacaloni, cameriere:2015} also implemented a frontier research at the edge of computer science and forensic medicine.
	This paper tries to move on the path of~\cite{Chen2019}, to implement a machine learning approach to forensic medicine and subject identification.
	Something similar has been done in~\cite{Ratha:2003} for what concerns fingerprint identification, however to authors' knowledge, this is the first application of object detection to forensic odontology.
	
	As discussed above, still a lot of work can be done on these subjects, not only to improve computational efficiency, but also to find new and interesting points of views.
	Indeed, working at the limit of different fields can yield new insights about both subjects, giving useful hints for a deeper and more complete understanding of all processes involved.
	%	
	
	%%%
	%%
	%
	\vskip 1cm
%\paragraph*{Acknowledgments} 
%	%
%	This work would not have been possible without the crucial contribution of dr. Gustavo de Felice.
%	Authors would like to thank him for the incredibly deep and useful discussions not only about forensic medicine.
%	Images used are also kindly provided by him that we thank for the courtesy.
	%
	%%%%%%%%%%%%%%%%%%%%%%%%%%%%%%%%%%%%%%%%%%%%%%%%%%%%%%%%%%%%%%%%%%%%%%%%%%%%%%%%
%%%%%%%%%%%%%%%%%%%%%%%%%%%%%%%%%%%%%%%%Appendices%%%%%%%%%%%%%%%%%%%%%%%%%%%%%%%%%%%%%%%%
	%%%%%%%%%%%%%%%%%%%%%%%%%%%%%%%%%%%%%%%%%%%%%%%%%%%%%%%%%%%%%%%%%%%%%%%%%%%%%%%%
	%
	%%
	%%%
	%%%%
	%%%
	%%
	%
	\appendix
	%
	%%
	%%%
	%%%
	%%
	%
	%Appendix 
	%%%%
	%%%
	%%
	%
\section{Working scheme of SIFT algorithm}
	\label{app:SIFT}
	In this appendix we face a detailed discussion about \textsc{sift} algorithm.
	We refer to the original paper~\cite{lowe1999sift} (that we review in this appendix) for further details.
	As said in the principal section~\ref{sec:SIFT}, images can be described by vectors of features. 
	These feature vectors do not only enjoy the nice property of being scale-invariant, but they are also invariant to translation, rotation, and illumination. In other words: everything a descriptor should be!

	As discussed, these descriptors are useful for matching objects are patches between images. 
	For example, consider creating a panorama. 
	Assuming each image has some overlapping parts, you need some way to align them so we can stitch them together. 
	If we have some points in each image that we know correspond, we can warp one of the images using a homography. 
	\textsc{sift} helps with automatically finding not only corresponding points in each image, but points that are also easy to match.
		\begin{figure}[!h]
			\includegraphics[width=0.9\linewidth]{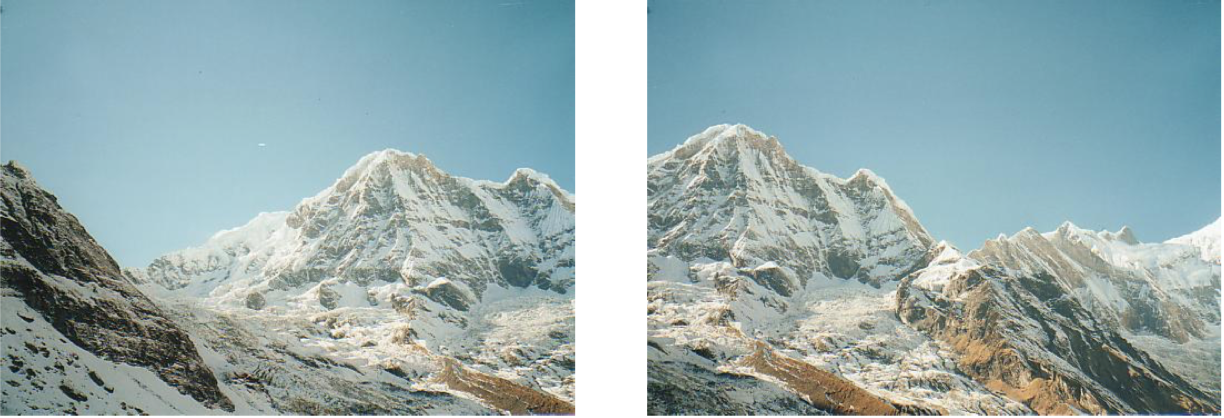}
			\caption{Two images with an overlapping region. 
				     The algorithm finds points to match the images in this region.}
		\label{fig:2-to-panorama}
		\end{figure}
		\begin{figure}[!h]
			\includegraphics[width=0.9\linewidth]{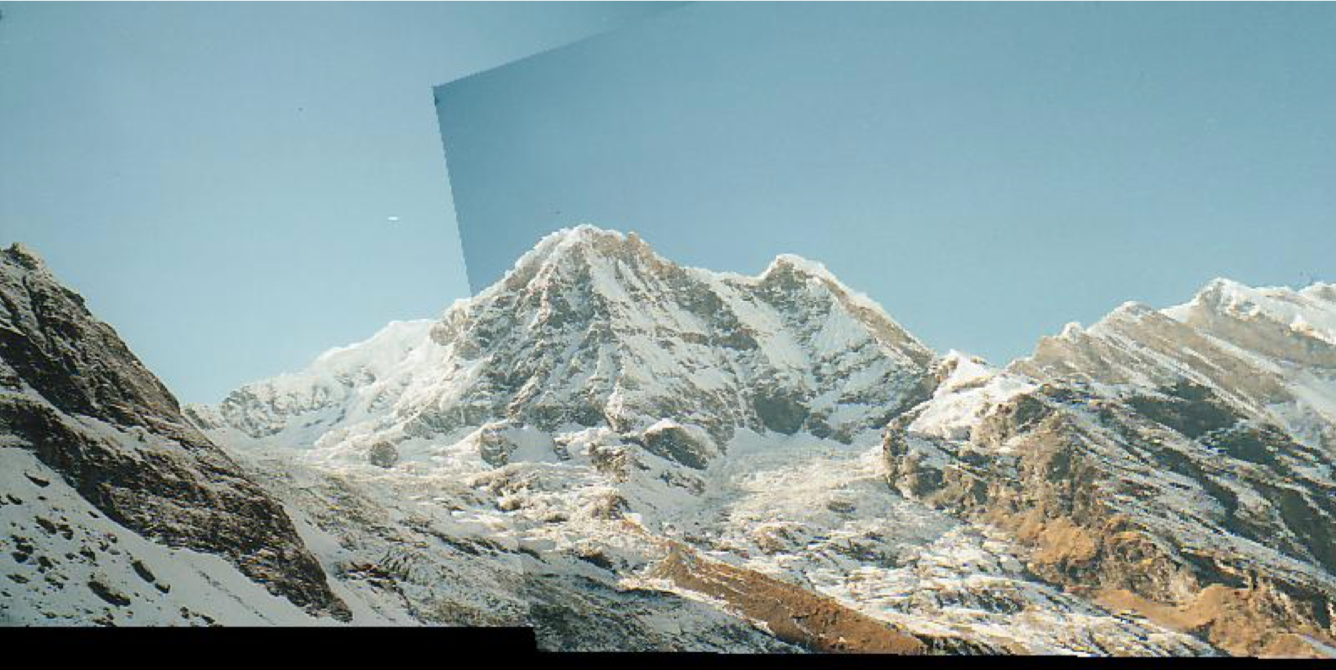}
			\caption{SIFT found keypoints on which we can match images.}
		\label{fig:panorama}
		\end{figure}
	One can find the many algorithm descriptions on the internet -- Wikipedia has a page dedicated to \href{https://en.wikipedia.org/wiki/Scale-invariant_feature_transform}{\textsc{SIFT}}.
	Here, we give a brief description focusing on the aspects useful for our purposes.
	First of all, we can split the algorithm in four main steps
		\begin{enumerate}
			\item Scale-space building and extrema detection
			\item Keypoint localisation
			\item Orientation assignement
			\item Local descriptors creation
		\end{enumerate}
	We are going to describe each of them to just have in mind how it works. .

	To begin, for any object in an image, interesting points on the object can be extracted to provide a ``feature description'' of the object. 
	This description, extracted from a given image, can then be used to identify the object when attempting to locate the object in a second image containing possibly many other objects. 
	As said, to perform reliable recognition, it is important that the features extracted from the training image be detectable even under changes in image scale, noise and illumination. 
	Such points usually lie on high-contrast regions of the image, such as object edges.

	The SIFT descriptor is based on image measurements in terms of receptive fields over which local scale invariant reference frames are established by local scale selection.

%%%
%%
%
\subsection{Constructing the scale space}
	Keypoints to identify are defined as extrema of a \emph{Gaussian difference} in a scale space defined over a series of smoothed and resampled images.

\begin{figure}[!h]
			\includegraphics[width=0.9\linewidth]{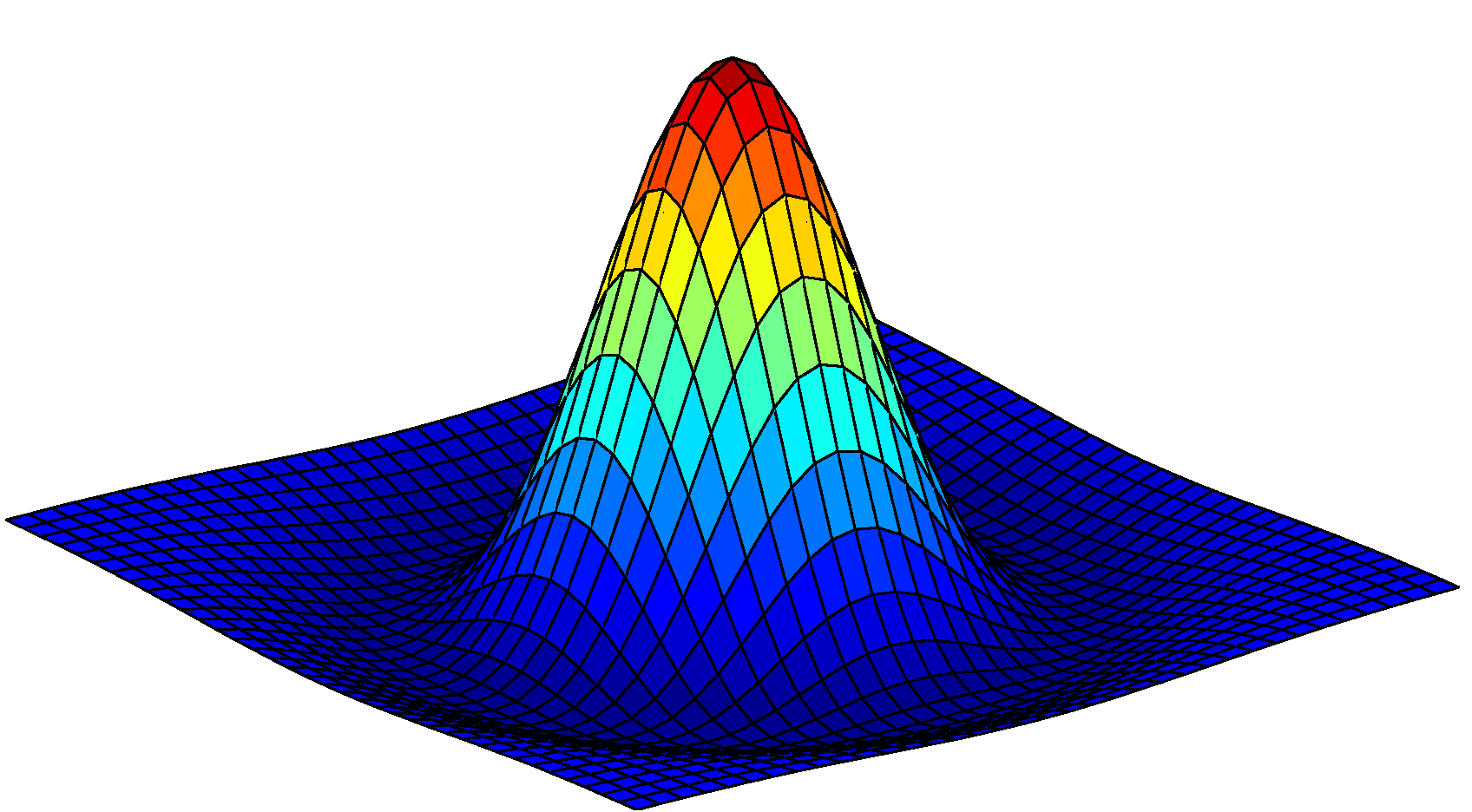}
			\caption{LoG filter is highly peaked at the center while becoming slightly negative and then zero at a distance from the center.}
		\label{fig:log}
		\end{figure}

	Hence, to begin we need to define a scale space and ensure that the keypoints we are going to select will be \emph{scale-independent}.
	In order to get rid of the noise of the image we apply a \emph{Gaussian blur}, while the characteristic scale of a feature can be detected by a scale-normalised Laplacian of Gaussian (\textbf{LoG}) filter.
	In a plot, a LoG filter looks like in figure~\ref{fig:log}.

	As one can observe, the typical shape of LoG filter is characterised by the standard deviation $\sigma$ of the Gaussian.

	The scale-normalisation for the LoG filter correspond to $\sigma^2 \mathrm{LoG}$ and it is used to correct the behaviour of the response of the LoG filter for a wider Gaussian that would be lower than for a smaller $\sigma$ Gaussian.

	The main issue with such a filter is that is expensive from a computationally point of view, this is due to the fact we have to calculate it to different scales, to make the procedure scale-invariant.
	Thankfully, even originally in the paper~\cite{lowe1999sift}, the authors of SIFT came up with a clever way to efficiently calculate the LoG at many scales.

	It turns out that the difference of two Gaussians (or \textbf{DoG}) with similar variance yields a filter that approximates the scale-normalized LoG very well:
		\begin{figure}[!h]
			\includegraphics[width=0.9\linewidth]{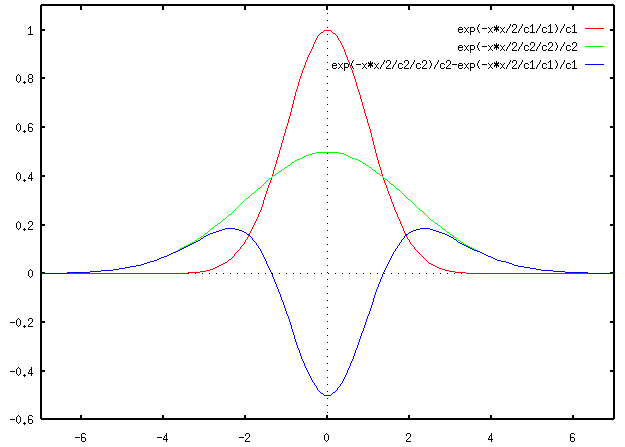}
			\caption{The difference of Gaussians approximates quite well the Laplacian.}
		\label{fig:dog}
		\end{figure}
	Thus, such approximation gives us an efficient way to estimate the LoG. 
	Now, we need to compute it at multiple scales. 
	SIFT uses a number of so-called \emph{octaves} to calculate the DoG. 
	The name might suggest that an octave means that eight images are computed. 
	However, an octave is actually a set of images were the blur of the last image is double the blur of the first image.
%		%
%		\begin{figure}[!h]
%			\includegraphics[width=0.9\linewidth]{Images/eiffel_blur.jpg}
%			\caption{Gaussian blur of an image.}
%		\label{fig:eiffel_blur}
%		\end{figure}
%		%

	All these filters and scales will multiply the number of images to consider -- or better, the number of versions of the same image. 
	At the end of the process we will end up with blur (Gaussian filter applied) images, created for multiple scales. 
	To create a new set of images of different scales, we will take the original image and reduce the scale by half. 
	For each new image, we will create different blur versions.% as in figure~\ref{fig:eiffel_blur}.

%	In figure~\ref{fig:eiffel_oct} one can find an example of octaves' creation making use of a picture of the Eiffel tower. 
	
	To create the octave, we first need to choose the number of images we want in each octave. 
	This is denoted by $s$. 
	Then $\sigma$ for the Gaussian filter is chosen to be $2^{1/s}$. 
	Since blur accumulates multiplicatively, when we blur the original image with this filter $s$ times, the result will have 
		\begin{equation*}
			\mathrm{blur} = 2 \times \mathrm{original\, blur}\, .
		\end{equation*}
	One detail from the Lowe's paper that is rarely seen mentioned is that in each octave, one actually needs to produce $s+3$ images (including the original one). 
	This is because when adjacent levels are subtracted to obtain the approximated LoG octave (\emph{i.e.} the DoG), we will get one less image than in the Gaussian octave -- see figure~\ref{fig:oct-diff}.
		\begin{figure}[!h]
			\includegraphics[width=0.9\linewidth]{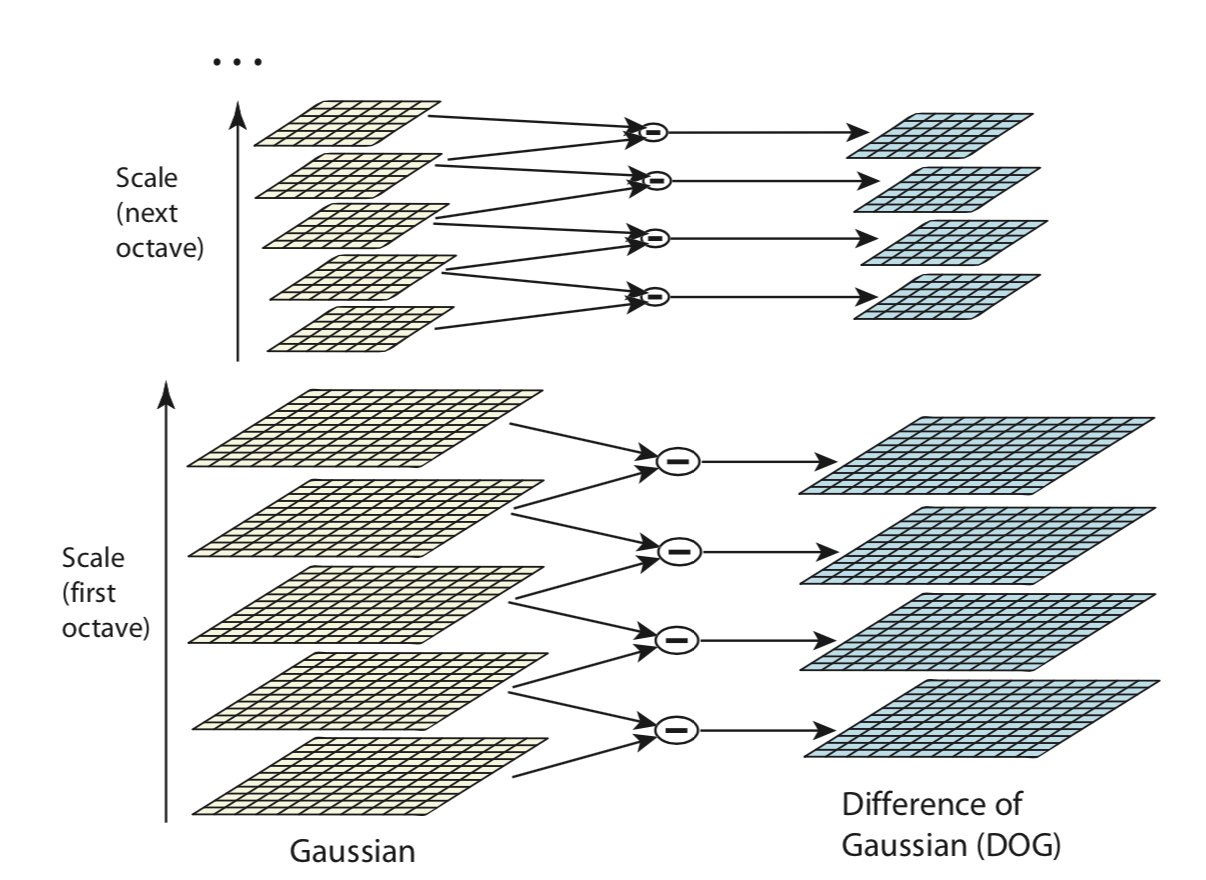}
			\caption{Difference of Gaussians.
				      Image from~\cite{lowe1999sift}.}
		\label{fig:oct-diff}
		\end{figure}
	Now we have $s+2$ images in the DoG octave. 
	However, later when we look for extrema in the DoG, we will look for the minimum or maximum of a neighbourhood specified by the current and adjacent levels.
	We will describe this later on, for the moment being, we have generated the Gaussian octave, we downsample the top level by two and use that as the bottom level for a new octave. 
	In~\cite{lowe1999sift}, author uses four octaves.
\paragraph*{Summary scheme}
	Just to sum up, here we collect the main points of the part one of the algorithm, that is the construction of scale-space or \emph{Gaussian pyramid}.
		\begin{itemize}
			\item[-] Given the original image, apply the blur filter to add a double the blur $s$ times.
			\item[-] Half the scale of the image to create different octaves.
			\item[-] Apply the DoG to get the feature enhanced version of the octave.
		\end{itemize}
		%
	%
%%%
%%
%
\subsection{Keypoint localisation}
	Once the scale space has been defined, we are ready to localise the keypoints to be used for feature matching.
	The idea is to identify extremal points (maxima and minima) for the feature enhanced images.

	To be concrete, we split this in two steps:
		\begin{itemize}
			\item[-] Find the extrema
			\item[-] Remove low contrast keypoints (also known under the name of \emph{keypoint selection})
		\end{itemize}
\subsubsection*{Extremal point scanning}
	We will not dig into details of extremisation algorithms to find maxima and minima. 
	We just give an heuristic insight.
	Conceptually, we explore the image space (\emph{i.e.} pixel by pixel) and compare each point value with its neighbouring pixels.
		\begin{figure}[!h]
			\includegraphics[width=0.9\linewidth]{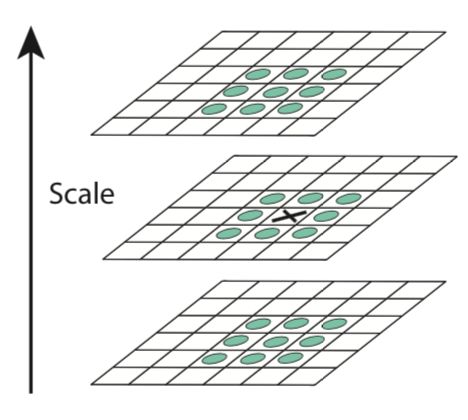}
			\caption{Extrema scanning in the octave space.
				      Image from~\cite{lowe1999sift}.}
		\label{fig:oct-diff}
		\end{figure}
	In other words, we scan over each scale-space DoG octave, $\mathcal{D}$, and include the center of each $3 \times 3 \times 3$ neighbourhood as a keypoint if it is the minimum or maximum value in neighbourhood.
	
	This is the reason the algorithm has generated $s+2$ levels in the DoG octave.
	One cannot scan over the points in the top or bottom level, but one still wants to get keypoints over a full octave of blur.

	Keypoints so selected are scale-invariant, however, they yield many poor choices and/or noisy, so in the next section we will throw out bad ones as well refine good ones.
\subsubsection*{Keypoint selection}
	The guide principle leading us to keypoint selection is 
		\begin{center}
			\begin{tcolorbox}[width=.9\linewidth, colframe = white]
				\emph{We want to eliminate the keypoints that have low contrast, or lie very close to the edge.}
			\end{tcolorbox}
		\end{center}

	This because low-contrast points are not robust to noise, while keypoints on edges should be discarded because their orientation is ambiguous, thus they will spoil rotational invariance of feature descriptors.
	
	The recipe to cook good keypoints goes through three steps:
		\begin{enumerate}
			\item Compute the subpixel location of each keypoint
			\item Throw out that keypoint if it is scale-space value at the subpixel is below a threshold.
			\item Eliminate keypoints on edges using the Hessian around each subpixel keypoint.
		\end{enumerate}
	In many images, the resolution is not fine enough to find stable keypoints, \emph{i.e.} in the same location in multiple images under multiple conditions. 
	Therefore, one can perform a second-order Taylor expansion of the DoG octave to further localise each keypoint. 
	Explicitly,
		\begin{equation}
			\mathcal{D} = \mathcal{D} + \partial_x \mathcal{D}^T + \frac{1}{2} \mathbf{x}^T \left(\partial^2_{\mathbf{x}} \mathcal{D}\right) \mathbf{x} \, .
		\end{equation}
	Here, $\mathbf{x}$ denotes the three-dimensional vector $[x, y, \sigma]$ corresponding to the pixel location of the candidate keypoint. 
	Taking the derivative of this equation with respect to $\mathbf{x}$ and setting it equal to zero yields the subpixel offset for the keypoint,
		\begin{equation}
			\overline{\mathbf{x}} = - \left(\partial^2_{\mathbf{x}} \mathcal{D}\right)^{-1} \partial_\mathbf{x} \mathcal{D} \, .
		\end{equation}
	This offset is added to the original keypoint location to achieve subpixel accuracy.

	At this stage, we have to deal with the low contrast keypoints.
	To evaluate if a given keypoint has low contrast, we perform again a Taylor expansion.

	Remind we do not just have keypoints, but subpixel offsets.
	The subpixel keypoint contrast can be calculated as,
		\begin{equation}
			\mathcal{D}(\overline{\mathbf{x}}) = \mathcal{D} + \frac{1}{2}\partial_\mathbf{x}\mathcal{D}^T \overline{\mathbf{x}}\, ,
		\end{equation}
	which is the subpixel offset added to the pixel-level location. 
	If the absolute value is below a fixed threshold, we reject the point. 
	We do this because we want to be sure that extrema are effectively extreme.

	Finally, as said we want to eliminate the contribution of the edge keypoints, because they will break rotational invariance of the descriptors.
	To do this, we use the Hessian calculated when computing the subpixel offset. 
	This process is very similar to finding corners using a \href{https://en.wikipedia.org/wiki/Harris_Corner_Detector}{Harris corner detector}.

	The Hessian has the following form,
		\begin{equation}
			\mathcal{H} = \begin{pmatrix} \mathcal{D}_{xx} & \mathcal{D}_{xy} \\ \mathcal{D}_{yx} & \mathcal{D}_{yy} \end{pmatrix}\, .
		\end{equation}

	To detect whether a point is on the edge, we need to \emph{diagonalise}, that is find eigenvalues and eigenvectors of such Hessian matrix.
	Roughly speaking and being schematic, if the eigenvalues of $\mathcal{H}$ are both large (with respect to some predetermined scale), the probability for the point to be on the edge is high. 
	We refer again to~\cite{lowe1999sift} for further details.
\subsubsection*{Orientation assignement}
	Getting the keypoints is only half the battle. 
	Now we have to obtain the actual descriptors. 
	But before doing so, we need to ensure another type of invariance: rotational.

	We have ensured \emph{translational invariance} thanks to the convolution of our filters over the image. 
	We also have \emph{scale invariance} because of our use of the scale-normalised LoG filter. 
	Now, to impose \emph{rotational invariance}, we assign the patch around each keypoint an orientation corresponding to its \emph{dominant gradient direction}. 

	Thus, to assign orientation, we take a patch around each keypoint thats size is proportional to the scale of that keypoint. See figure~\ref{fig:num-grad}.
		\begin{figure}[!h]
		\centering
			\begin{tikzpicture}
 				\matrix [matrix of math nodes,
    					nodes={rectangle, draw, minimum size=1cm},
  					] (M){
						35 & 40 & 41 & 45 & 50 \\
						40 & 40 & 42 & 46 & 52 \\
						42 & 46 & 50 & 55 & 55 \\
						48 & 52 & 56 & 58 & 60 \\
						56 & 60 & 65 & 70 & 75 \\};
					
				\draw[line width=0.50mm, green] (M-2-1.north east) -- (M-2-5.north west) -- (M-4-5.south west) -- (M-4-1.south east) -- cycle;
				\draw[line width=0.50mm, red] (M-3-2.north east) -- (M-3-4.north west) -- (M-3-4.south west) -- (M-3-2.south east) -- cycle;
				\draw[->, line width=1mm, blue] ([yshift=4pt]M-3-4.south) -- ([yshift=4pt]M-3-2.south);
				\draw[->, line width=1mm, blue] ([xshift=-4pt]M-4-4.west) -- ([xshift=-4pt]M-2-4.west);
			\end{tikzpicture}
			\caption{Example of an array of pixels and numerical gradient in its central point.}
		\label{fig:num-grad}
		\end{figure}

	Given an array of pixels, one can calculate the following quantities,
		\begin{equation*}
			\begin{aligned}
				r = \sqrt{\left(\partial_x f\right)^2 + \left(\partial_y f\right)^2}		& & \mbox{\textbf{magnitude}}\\ 
				\phi = \arctan{\left(\partial_x f / \partial_y f\right)}			& &\mbox{\textbf{orientation}}
			\end{aligned}
		\end{equation*} 
	These are functions of the gradient in each point.
	To summarise, we can state the magnitude represents the intensity of the pixel and the orientation gives the direction for the same.

	We can now create a histogram given that we have calculated these magnitude and orientation values for the whole pixel space.

	The histogram is created on orientation value (the gradient is specified in polar coordinates) and has $36$ bins (each bin has a width of $10$ degrees). 
	When the magnitude and angle of the gradient at a pixel are calculated, the corresponding bin in our histogram grows by the gradient magnitude weighted by the Gaussian window.
	Once we have our histogram, we assign that keypoint the orientation of the maximal histogram bin.
\subsubsection*{Local descriptors}
	Finally we got at the final step of SIFT.
	The previous step gave us a set of stable keypoints, which are also scale-invariant and rotation-invariant.
	Now, we can create the local descriptors for each keypoint. 
	We will make use of points in the neighbourhood of each keypoints to characterise it completely.
	The keypoint endowed with these local properties is called a \emph{descriptor}.

	A side effect of this procedure is that, since we use the surrounding pixels, the descriptors will be partially invariant to illumination or brightness of the images.

	We will first take a $16 \times 16$ neighbourhood around the keypoint. 
	This $16 \times 16$ block is further divided into $4 \times 4$ sub-blocks and for each of these sub-blocks, we generate an $8$-bin histogram using magnitude and orientation as described above.
	Finally, all of these histograms are concatenated into a $4\times 4 \times 8 = 128$ element-long feature vector.

	This final feature vector is then normalised, thresholded, and renormalised to try and ensure invariance to minor lighting changes.
		\begin{figure}
			\includegraphics[width=0.9\linewidth]{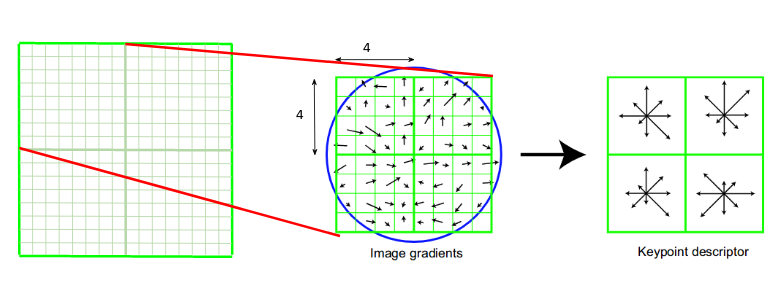}
			\caption{Each of these arrows represents the $8$ bins and the length of the arrows defines the magnitude. 
					So, we will have a total of $128$ bin values for every keypoint.}
		\label{fig:descr}
		\end{figure}
	To finally summarise, we can give the definition of a local descriptor.
		\begin{center}
			\begin{tcolorbox}[width=.9\linewidth, colframe = white]
				\emph{A local descriptor is a set of features, encoded by a feature vector, describing magnitude and orientation of neighbourhood for each keypoint.}
			\end{tcolorbox}
		\end{center}
	This leads us to the end of this long appendix, quite technical but describing the working scheme of the algorithm.
	%
	%%%
	%%
	%
%	%Appendix 
%	%%%%
%	%%%
%	%%
%	%
%\section{Another useful appendix}
%	%	
%	
%	%	
	%%%
	%%
	%
	%%%%%%%%%%%%%%Appendices%%%%%%%%%%%%%%%%%%%
	%
	%%
	%%%
%
	%%%
% 	%%
% 	%
% 	%
% 	
%%%%%Bibliography%%%%
\onecolumngrid
   	\bibliographystyle{unsrtsiam2}
	\bibliography{Bibliography}
	\nocite{*}
%%%%%%%%%%%%%%%

\end{document}